\documentclass[runningheads]{llncs}
\usepackage{graphicx}
\usepackage{amsmath} 
\usepackage[dvipsnames]{xcolor} 
\usepackage{cite} 
\usepackage{hyperref} 
\usepackage{booktabs} 
\usepackage{multirow} 
\usepackage{lscape} 
\usepackage{amsfonts} 
\begin{document}
%
\title{Explainable AI-Based Interface System for 
Weather Forecasting Model 
\thanks{Supported by the Korean Institute of Information \& Communications Technology Planning \& Evaluation (IITP) and the Korean Ministry of Science and ICT(MSIT) under grant agreement No. 2019-0-00075 (Artificial Intelligence Graduate School Program (KAIST)) and No. 2022-0-00984 (Development of Plug-and-Play Explainable Artificial Intelligence Method), and from the Korea Meteorological Administration (KMA) and Korean National Institute of Meteorological Sciences (NIMS) under grant agreement No. KMA2021-00123 (Developing Intelligent Assistant Technology and Its Application for Weather Forecasting Process).}}
\titlerunning{XAI-Based Interface System for Weather Forecasting Model}
%
\author{Soyeon Kim \inst{1}\orcidID{0009-0001-5037-0902} \and
        Junho Choi \inst{1}\orcidID{0000-0002-7800-6950} \and
        Yeji Choi \inst{2}\orcidID{0000-0002-8212-1126} \and
        Subeen Lee \inst{1}\orcidID{0009-0001-7996-4114} \and
        Artyom Stitsyuk \inst{1}\orcidID{0009-0005-6446-137X} \and
        Minkyoung Park \inst{1}\orcidID{0009-0004-9420-4406} \and
        Seongyeop Jeong \inst{1}\orcidID{0009-0008-8480-8117} \and
        Youhyun Baek \inst{3}\orcidID{0000-0001-6362-4353} \and
        Jaesik Choi \inst{1, 4}\orcidID{0000-0002-4663-3263}}




%
\authorrunning{Kim et al.}
%
\institute{Korea Advanced Institute of Science and Technology(KAIST), Daejeon, Korea \\
\email{soyeon.k, junho.choi, forestsoop, stitsyuk, jrneomy, seongyeop.jeong@kaist.ac.kr} \and
SI-Analytics, Daejeon, Korea \\
\email{yejichoi@si-analytics.ai} \and
National Institute of Meteorological Sciences(NIMS), Jeju, 63568, Korea
\email{yhbaek88@korea.kr} \and
INEEJI, Gyeonggi, Korea \\
\email{jaesik.choi@kaist.ac.kr}}
\maketitle              
\begin{abstract}
Machine learning (ML) is becoming increasingly popular in meteorological decision-making.
Although the literature on explainable artificial intelligence (XAI) is growing steadily, user-centered XAI studies have not extend to this domain yet. 
This study defines three requirements for explanations of black-box models in meteorology through user studies: statistical model performance for different rainfall scenarios to identify model bias, model reasoning, and the confidence of model outputs. 
Appropriate XAI methods are mapped to each requirement, and the generated explanations are tested quantitatively and qualitatively. 
An XAI interface system is designed based on user feedback.
The results indicate that the explanations increase decision utility and user trust. 
Users prefer intuitive explanations over those based on XAI algorithms even for potentially easy-to-recognize examples. 
These findings can provide evidence for future research on user-centered XAI algorithms, as well as a basis to improve the usability of AI systems in practice.   

\end{abstract}

\begin{figure}[t!]
\includegraphics[width=\textwidth]{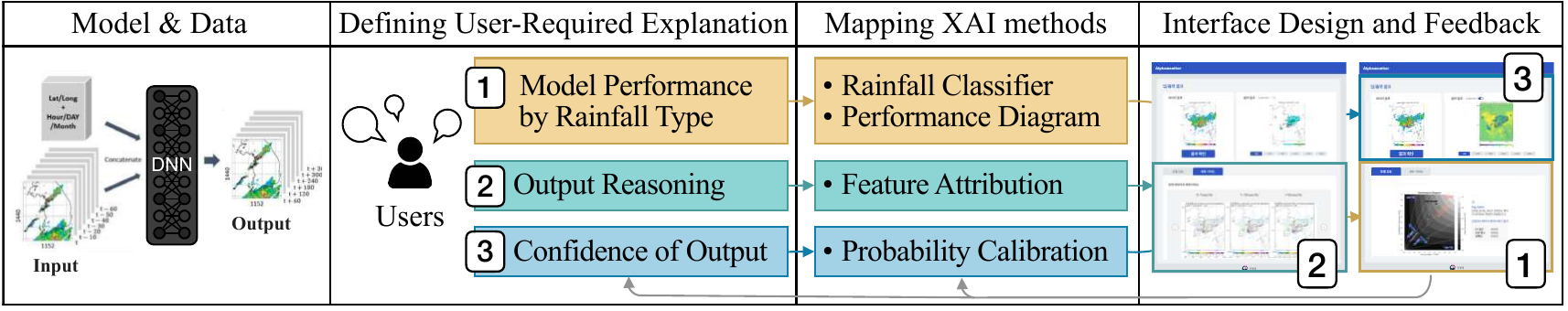}
\caption{Workflow for developing a user-centered explainable artificial intelligence (XAI) interface system. The system is developed based on the procedures established in the previous literature \cite{liao2021question, liao2020questioning}. The scope of explanations is defined based on the requirements set by the practitioners; appropriate XAI algorithms are selected based on the defined scope; and the interface is designed with user feedback} \label{main}
\end{figure}

\section{Introduction}
Weather prediction has always been an integral part of human society due to its significant socioeconomic impact, influencing various aspects such as agricultural productivity, industrial output, labor efficiency, energy demand, public health, conflicts, economic growth, \cite{dell2014we} as well as ecosystems and their ecosystem services \cite{van2019impacts}.
With the increasing volatility of meteorological patterns caused by the climate crisis, economic losses from extreme weather events are on a rapid incline \cite{zhongming2021atlas,mizutori2017economic}. 
Accurate weather forecasting is crucial for mitigating the effects of these scenarios.

Operational weather forecasting is conventionally performed through Numerical Weather Prediction (NWP), a process of simulating future weather patterns using a comprehensive set of equations that describe the physical dynamics of the atmosphere \cite{kalnay2003atmospheric}. 
Although it has a long history and sees use even today, NWP faces several challenges such as high computational costs and sensitivity to the derived initial conditions \cite{ren2021deep}. 
Data-driven deep learning models for weather prediction are seen as a potential alternative, being able to exploit the growing availability of weather data and make predictions for a fraction of the cost of operating NWP models \cite{rasp2021data}.

One issue faced by practitioners in producing weather forecasts is the vast amount of documents required to produce the forecasts.
For example, Korea Meteorological Agency (KMA) creates 2.2TB worth of data daily on average for weather forecasts  \cite{kma2020haneulsarang}.
The sheer size of the data can be extremely burdensome for the forecasters, who not only have limited time when making short-term forecasts and associated decision-making, but also need to continuously monitor the occurrence of sudden extreme weather patterns.
One of the reasons for requiring large data lies with the difficulty in accurate prediction of rainfall.
If the accuracy of rainfall prediction can be improved through the use of deep learning, it could reduce some of the burden placed on the forecasters so that their efforts could be invested elsewhere. 

A key issue preventing the use of deep learning models in operational forecasting is their lack of interpretability \cite{ren2021deep}. 
While the state-of-the-art models \cite{espeholt2022deep,ravuri2021skilful,kim2020precipitation,sonderby2020metnet} may make accurate predictions, they tend to be black boxes -- a user cannot determine how the models infer these outcomes. 
A forecaster would not be able to accept predictions without sufficient justifications due to the high stakes associated with wrong predictions. 
The extensive array of techniques in the field of explainable artificial intelligence (XAI) can help meet these requirements \cite{gilpin2018explaining,schwalbe2023comprehensive,adadi2018peeking}; unfortunately, the sheer number of available techniques makes it difficult to determine which methods should be used. 
One potential approach of filtering the appropriate techniques is to center the explanations around its intended audience. 
An appropriate explanation is dependent on the task performed by a model and the audience of the explanation \cite{murdoch2019definitions,liao2021human}. 
Therefore, an explanation system should be centered around its users, regardless of domain. 
A recent study of the user-explained AI (UXAI) \cite{chaput2021explanation} even claims that users may not be satisfied by an explanation that has considered the users in its design if it has not been made \textit{with} the users.
Despite the increasing interest in both user-centric \cite{liao2021human} and regular XAI in the meteorological domain \cite{bacsaugaouglu2022review,mcgovern2022we,mcgovern2014enhancing}, there seems to be a distinct lack of user-centered XAI studies in meteorology.
This paper attempts to fill this gap by following a user-centered XAI framework to create a prototype system that explains a precipitation prediction AI model. 
Specifically, the paper follows the process described by \cite{liao2020questioning} and \cite{liao2021question}: (a) the scope of explanations is defined through an XAI question bank, which divides the typical questions that could be asked by a user into several major categories, (b) appropriate XAI methods are selected based on the categories that the questions belong to, and (c) an interface system is designed based on user input and feedback to express the explanations.

The main contributions of this paper are as follows:
\begin{itemize}
    \item Demonstrates the procedures of the user-centric XAI development framework from an operational perspective.
    \item Creates a user-experience-based prototype of the XAI system in the meteorological domain.
    \item Analyzes the available XAI methods and discusses their practical limitations.
\end{itemize}

The eventual objective of our work is to provide accurate and trustworthy information required by the user as an end-product of a single map, reducing the procedural burden shouldered by the forecasters in the current system.

\section{Materials}

\begin{figure}[t!]
\includegraphics[width=\textwidth]{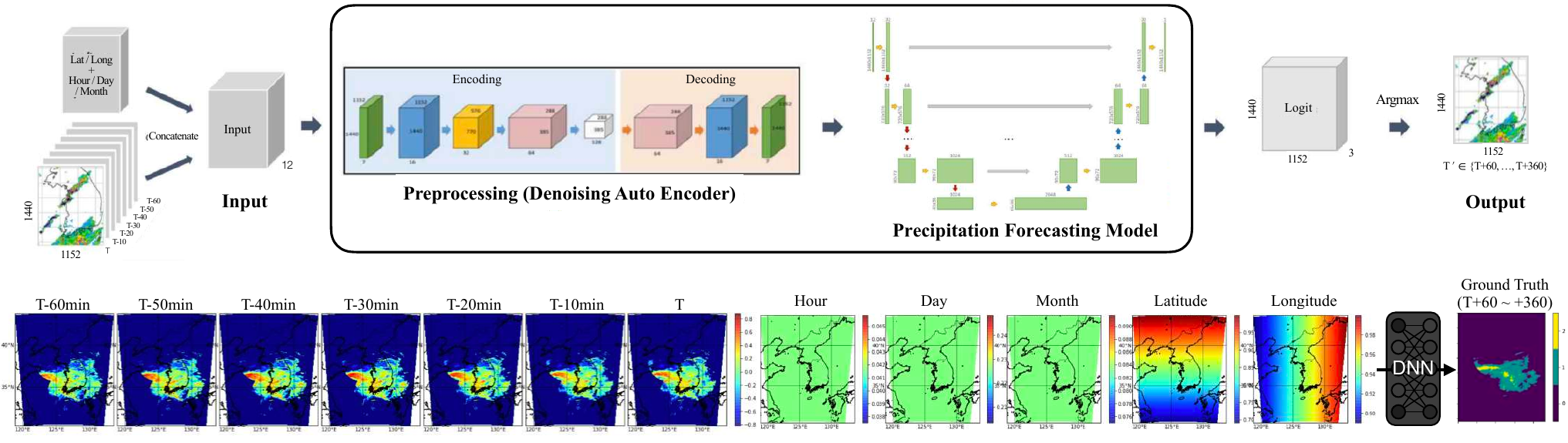}
\caption{The target precipitation forecasting model and data. The data consists of radar hybrid scan reflectivity.} 
\label{model}
\end{figure}

\begin{figure}[b!]
\includegraphics[width=\textwidth]{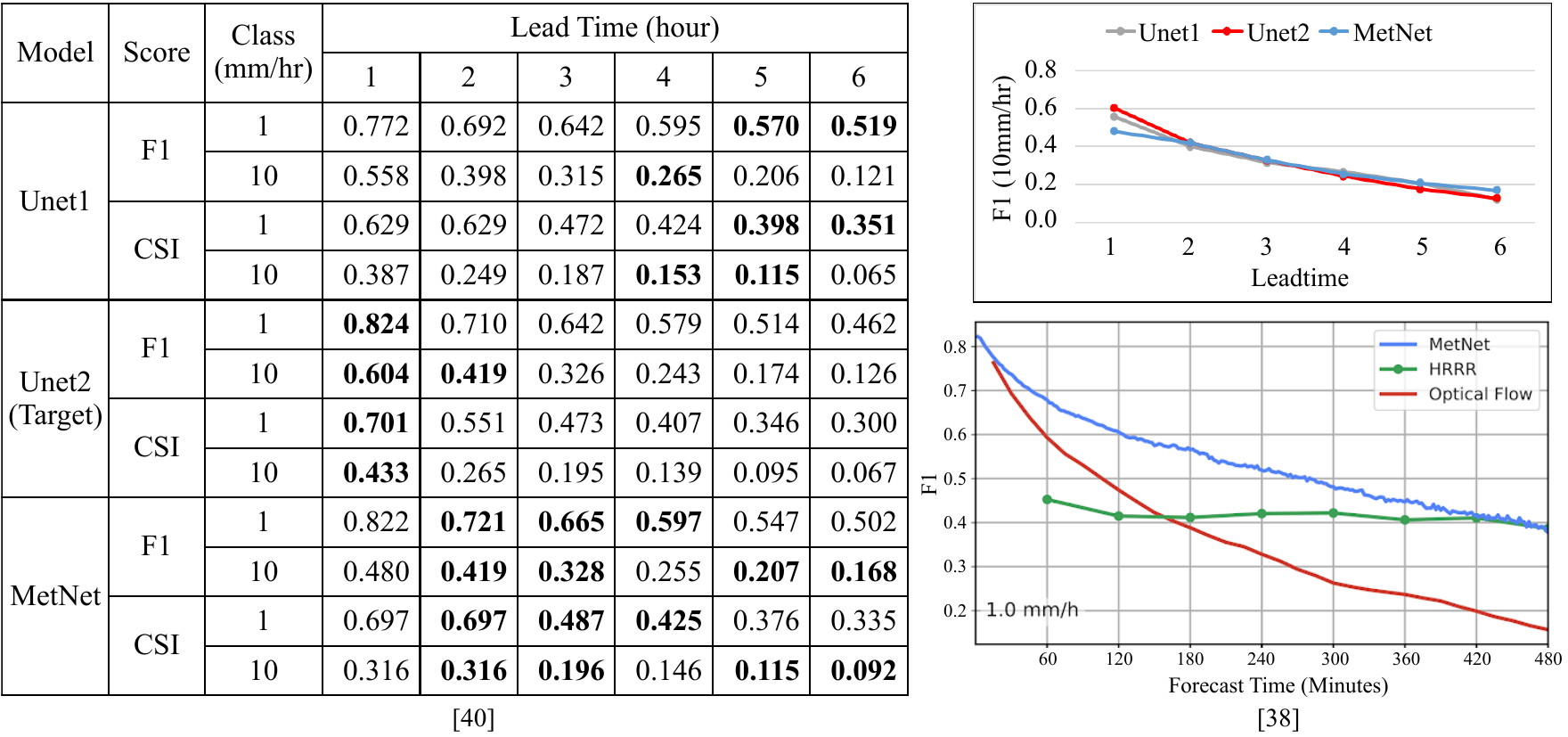}
\caption{The performance of the target model. UNet1 and UNet2 built by NIMS are comparable to MetNet \cite{sonderby2020metnet} and HRRR numerical model for very short-term predictions. Reproduced from \cite{sonderby2020metnet} and \cite{yun2021development}.} \label{performance}
\end{figure}

\subsection{Model and Data}
The aim of this study is to design a user-centered interface system for explaining UNet2, a UNet-based model (an unpublished variant of DeepRaNE \cite{ko2022effective}) developed by the National Institute of Meteorological Sciences (NIMS) for 2020 radar synthesis data for very short-term rainfall intensity prediction (Figure \ref{model}).
UNet2 consists of a denoising autoencoder followed by a convolutional neural network-based U-Net architecture and addresses a segmentation task of predicting three rainfall intensity intervals (no rain 0-1 mm/hr, light rain 1-10 mm/hr, and heavy rain 10 mm/hr over) between 1 and 6 hours in the future at 1-hour intervals.
The class intervals have been established by domain experts.
The input data consists of seven radar data sequence at ten minutes intervals, two spatial features for longitude and latitude, and three temporal features for year, month, and day of the current date. The data are concatenated into 12 channels following an early fusion scheme. 
The performance of UNet2 is comparable to the MetNet \cite{sonderby2020metnet} and HRRR numerical models for very short-term predictions(Figure \ref{performance}). 
In particular, for rainfall prediction with a one-hour lead time and rainfall rates of 1-10mm/hr, UNet2 and MetNet achieve F1 scores of 0.824 and 0.822, respectively.
For heavy rainfall rates over 10mm/hr, UNet2 and MetNet have F1 scores of 0.604 and 0.480, respectively.

\section{Methods}
\subsection{User Requirements of Explanation}\label{exp0}
This study has been performed with discussions from sixteen online meetings with NIMS, as well as three in-person external advisories from domain experts from 27 April 2022 to 12 April 2023.

\subsubsection{User Study.}
An XAI question bank \cite{liao2021question, liao2020questioning} is utilized in the early phase of interviews to brainstorm the desired explanations from AI systems. 
Based on the discussion, the user requirements can be stated as follows. 
First, forecasters are interested in the consistency of the model inferences in various rainfall situations. 
If systematic biases for each rainfall type are provided, it can help the forecasters decide whether to use the model in practice.
Second, forecasters consider the movement, growth, and dissipation of the convection cell as key factors for predicting the change of very short-term precipitation clouds around a 6-hour scale.
In particular, they would like to identify the precursors to pinpoint the seeds that are the most susceptible to convective system development.
Through the precursors, the users can indicate the locations that require more intensive monitoring.
Finally, the users are interested in the local reliability of the predictions.
For the rest of this study, these three requirements are referred to as model performance explanation by rainfall type, output reasoning explanation, and confidence of output explanation, respectively.

\subsubsection{Mapping XAI methods.}
Appropriate XAI methods are selected to address each need.
First, a rainfall type classifier is combined with performance diagram for each rainfall type for generating a model performance explanation (Section \ref{exp1}).
Second, feature attribution is used for output reasoning explanation since the associated techniques can evaluate the contributions of the input features for generating the predictions (Section \ref{exp2}).
Lastly, a probability calibration technique is adopted for model confidence explanation (Section \ref{exp3}).

\subsection{Explanation 1: Model Performance by Rainfall Types} \label{exp1}
\subsubsection{Rainfall Type Classifier.}
For this explanation, an input sample is assigned to a rainfall category using a deep learning classifier; then, the model's predictive performance for the corresponding rainfall type is analyzed. 
This setup allows for a comparison of model performance over different rainfall scenarios. 

The rainfall type classifier is built by fine-tuning the parameters from the pre-trained encoder of the target model. 
Self-organizing map (SOM)-based rainfall type classification data and its quantitative labels provided by NIMS based on the characteristics of the Korean Peninsula have been used for the experiment.
The five rainfall types are monsoon front (southern region), monsoon front (central region), isolated thunderstorm,  extratropical cyclone (east coast), and extratropical cyclone (inland). 
These precipitation types are often used by forecasters in practice.
29, 280, 53, 43, and 24 samples are used for each of the five types of rainfall in 2020.
Additionally, 218 cases are sampled in equal intervals for the no-rain type. 
The dataset is split into three portions: 60\% for training, 20\% for validation, and another 20\% for testing.
For the training dataset, a sampler that follows a multinomial distribution using the probability parameter as the inverse of the number of samples of each class in the dataset is used to solve the class imbalance problem.
The classifier is optimized using the Adam solver with a learning rate of 1e-6 and weight decay of 1e-8. 
Additionally, the weighted cross-entropy loss is adopted to account for the classes with deficient samples.
The classifier performance shows an accuracy of 93.07\%.

\begin{figure}[t!]
\includegraphics[width=\textwidth]{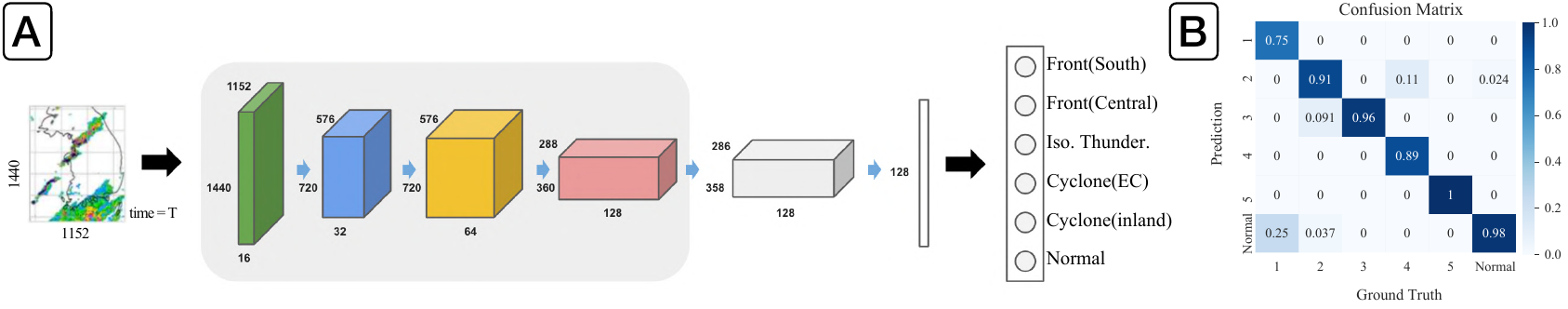}
\caption{The structure of the precipitation classifier (A) and the resulting confusion matrix (B). The rainfall types are based on a SOM-based weather classification study (an unpublished result of \cite{shin2022classification} with the same research procedure on a specific region).} \label{accuracy}
\end{figure}
\vspace{-0.1in}

\subsubsection{Performance Diagram.}
\begin{figure}[t!]
\includegraphics[width=\textwidth]{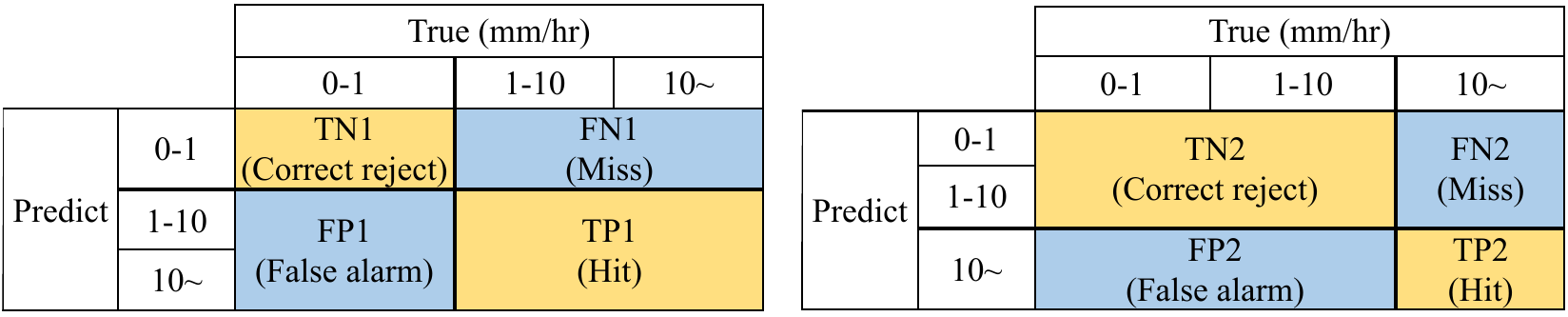}
\caption{Confusion matrices to calculate performance metrics on the imbalanced data.} \label{confusion}
\end{figure}

\begin{figure}[t]
\includegraphics[width=\textwidth]{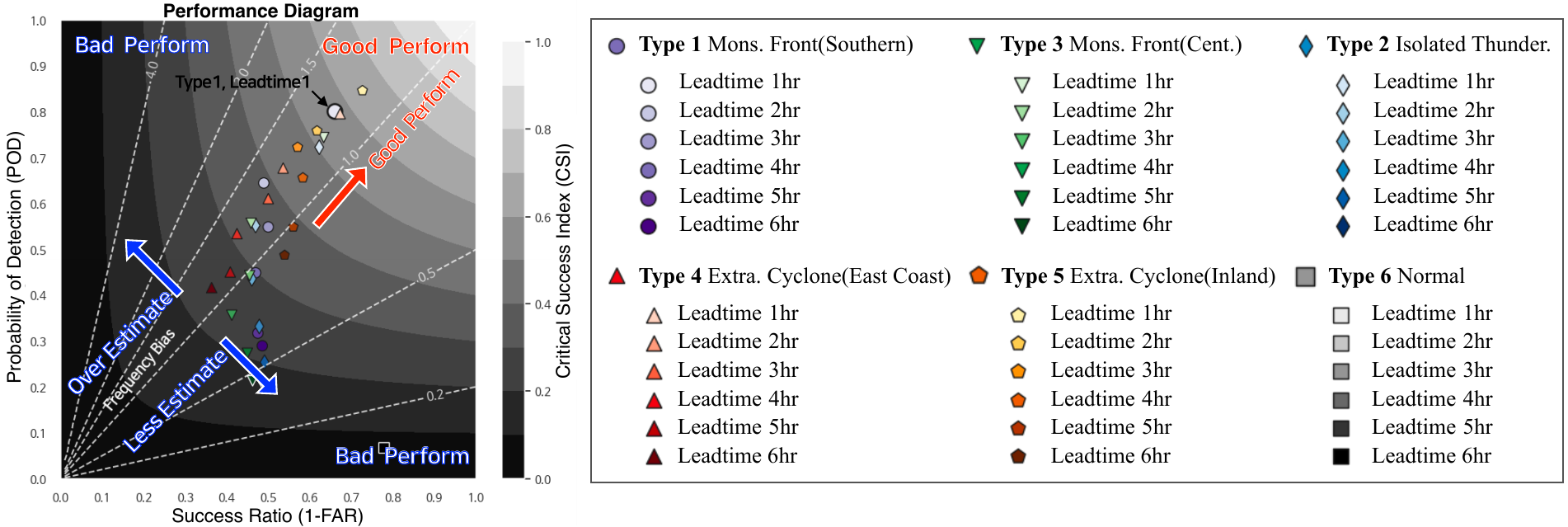}
\caption{Performance diagram. The diagram helps visualize the overall performance of bias, CSI, POD, and success ratio in a single chart.}
\label{performancediagram}
\end{figure}

The performance diagram is a method of visualizing the overall performance of a model \cite{roebber2009visualizing} and can express important model evaluation indicators in the meteorological domain such as bias, critical success index (CSI), probability of detection (POD), and success ratio in a single chart(Figure \ref{performancediagram}. 
To alleviate the problem of imbalanced rainfall intensities, where the rainfall amounts of interest infrequently occur in the real world, the metrics are computed for the light rainfall intensity and more (1 mm/hr over) and the heavy rainfall intensity (10 mm/hr over) as shown in Figure \ref{confusion} and are averaged.
Formally, 

\vspace{-0.1in}
\begin{equation}\label{pod}
\operatorname{Modified POD}=\frac{1}{2}\left( \right.
\frac{\mathrm{Hit}_{\text {1 (mm/hr) over }}}{\mathrm{Hit}_{ \text {1 over }}+ \mathrm{Miss}_{\text {1 over }}}
+\frac{\mathrm{Hit}_{\text {10 over }}}{\mathrm{Hit}_{\text {10 over }}+\mathrm{Miss}_{\text {10 over }}}\left. \right)
\end{equation}
\vspace{-0.1in}

\begin{equation}\label{far}
\begin{split}
\operatorname{Modified FAR}=\frac{1}{2}\left( \right.
& \frac{\mathrm{False Alarm}_{\text {1 (mm/hr) over }}}{\mathrm{False Alarm}_{ \text {1 over }}+ \mathrm{Hit}_{\text {1 over }}} \\
& +\frac{\mathrm{False Alarm}_{\text {10 over }}}{\mathrm{False Alarm}_{\text {10 over }}+\mathrm{Hit}_{\text {10 over }}}\left. \right)
\end{split}
\end{equation}
\vspace{-0.1in}

\begin{equation}\label{f1}
\begin{split}
\operatorname{Modified F1}=\frac{1}{2}\left( \right.
&\frac{\mathrm{Hit}_{\text {1 (mm/hr) over }}}{\mathrm{Hit}_{ \text {1 over }}+ \frac{1}{2}(\mathrm{Miss}_{\text {1 over }} + \mathrm{False Alarm}_{\text {1 over }})} \\
&+\frac{\mathrm{Hit}_{\text {10 over }}}{\mathrm{Hit}_{\text {10 over }}+\frac{1}{2}(\mathrm{Miss}_{\text {10 over }}+ \mathrm{False Alarm}_{\text {10 over }})}\left. \right)
\end{split}
\end{equation}

As results in Figure \ref{performancediagram}, the performance diagram shows that for a lead time of 1 hour, the model has the best performance for rainfall type 5 - inland extratropical cyclone. 
The model is a little overestimated overall, but less estimated on the long lead times.
The worst performance arises in the type of normal weather at the lead time of 6 hours. The low POD suggests that the model fails to predict the real rainfall at this lead time.

\subsection{Explanation 2: Output Reasoning}\label{exp2}

\begin{figure}[t!]
\includegraphics[width=\textwidth]{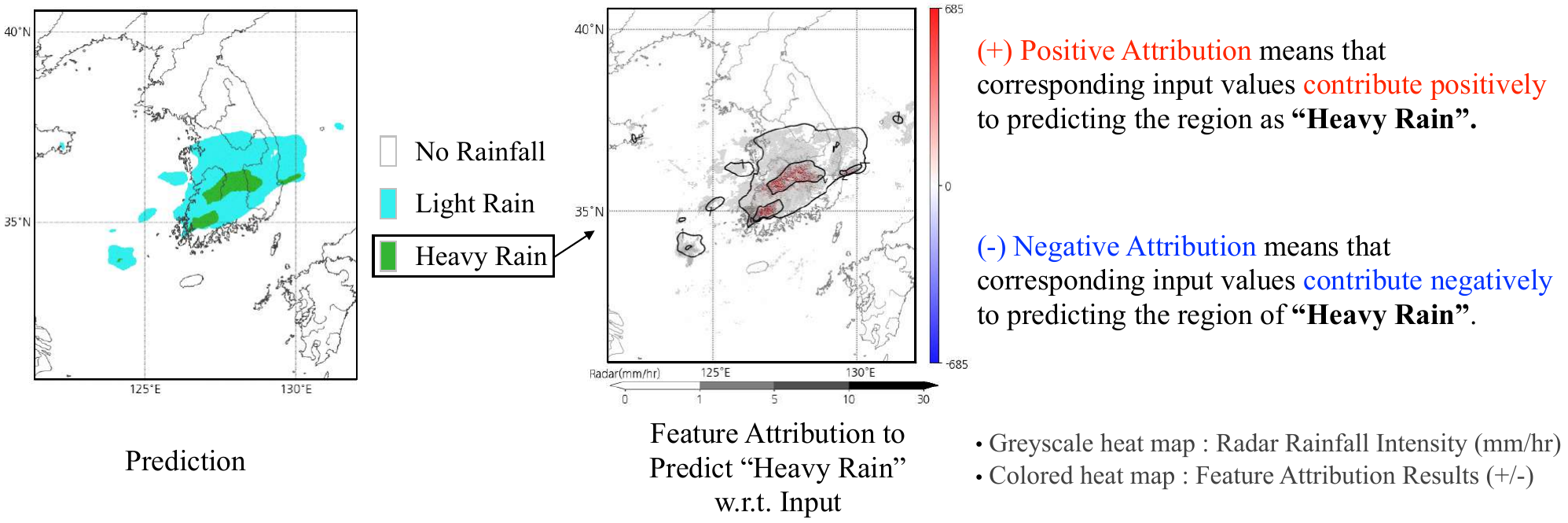}
\caption{Feature attribution. The heatmap describes the location and the degree of relevance of the inputs as the cause of the trained model prediction.}
\label{fa}
\end{figure}

Feature attribution methods analyze the contribution of the inputs for a model's prediction.
As shown in Figure \ref{fa} feature attribution methods allow users to investigate the reason why the model infers the development or the dissipation of a rain cell one hour later from the radar input.
There are many feature attribution methods available; even a list of some of the more prevalent methods (\textit{Saliency Maps} \cite{simonyan2013deep}, \textit{Integrated Gradients} \cite{sundararajan2017axiomatic}, \textit{GuidedGrad-CAM} \cite{selvaraju2017grad} and \textit{Layer-Wise Relevance Propagation (LRP)} \cite{bach2015pixel} to name a few) can be extensive. 
This study selects the attribution method by quantitatively evaluating the completeness of the generated attributions following the incremental deletion criterion \cite{nauta2022anecdotal, samek2016evaluating}: the predictive performance of the model should decrease as the inputs are removed sequentially based on their importance, with the speed of decline faster at the initial stages of removal compared to the latter stages. 
After selecting a method, sample cases are analyzed by domain experts to evaluate user opinions on the generated results.

\begin{figure}[b!]
\centering
\includegraphics[width=0.6\textwidth]{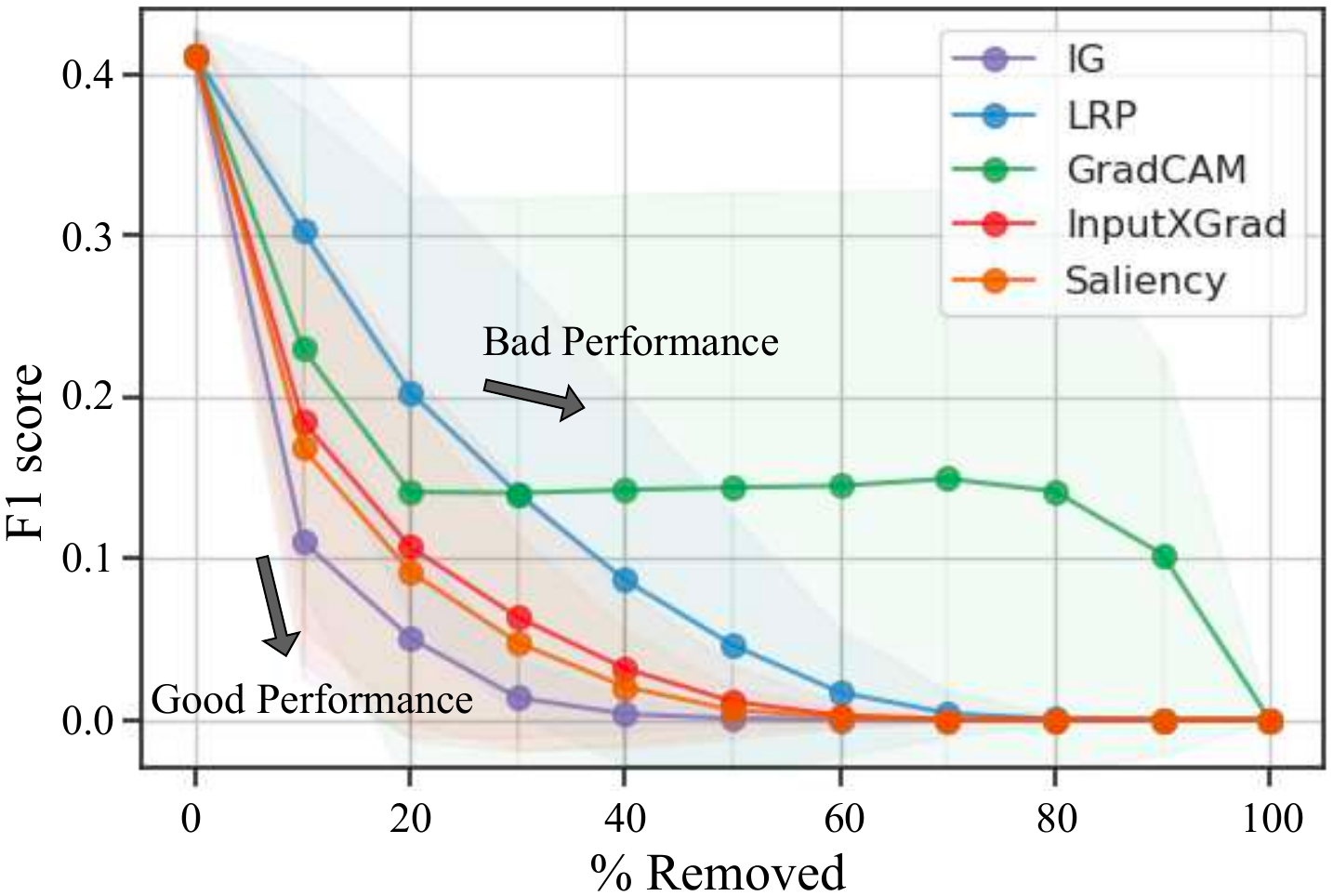}
\caption{Quantitative evaluation of the output reasoning explanation from different feature attribution methods.}
\label{fa_quantity}
\end{figure}

\subsubsection{Quantitative Evaluation with Incremental Deletion.}

To quantitatively compare how well the feature importance maps from different methods reflect the true relative contributions of the features to the model predictions, the level of performance reduction is evaluated after eliminating the Top K\% region of the input in the order of attribution value. 
A steeper decrease in performance implies greater fidelity. 
As shown in Figure \ref{fa_quantity}, the integrated gradient method outperforms the other methods.

\begin{figure}[t!]
\includegraphics[width=\textwidth]{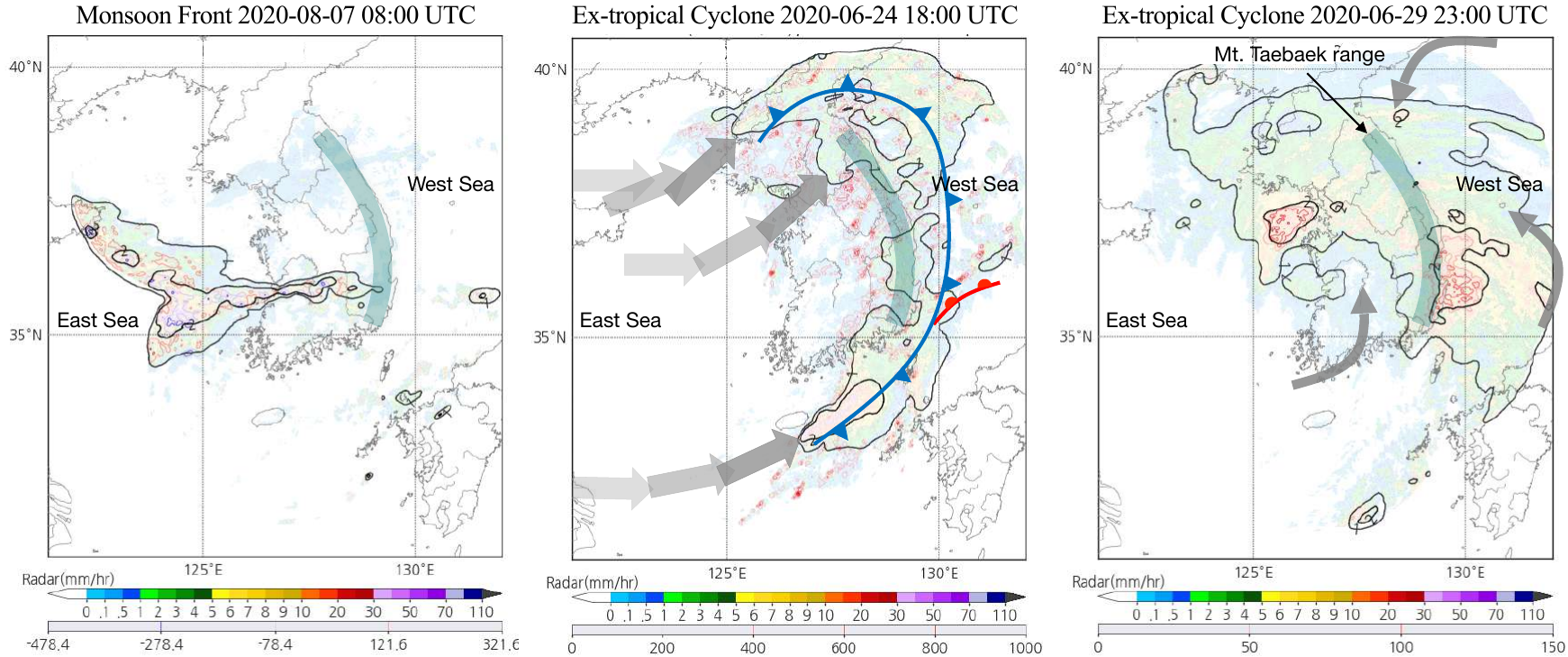}
\caption{Anecdotal evidence based on domain expert's case analysis of monsoon front and extratropical cyclones.}
\label{fa_quality}
\end{figure}

\begin{figure}[b!]
\includegraphics[width=\textwidth]{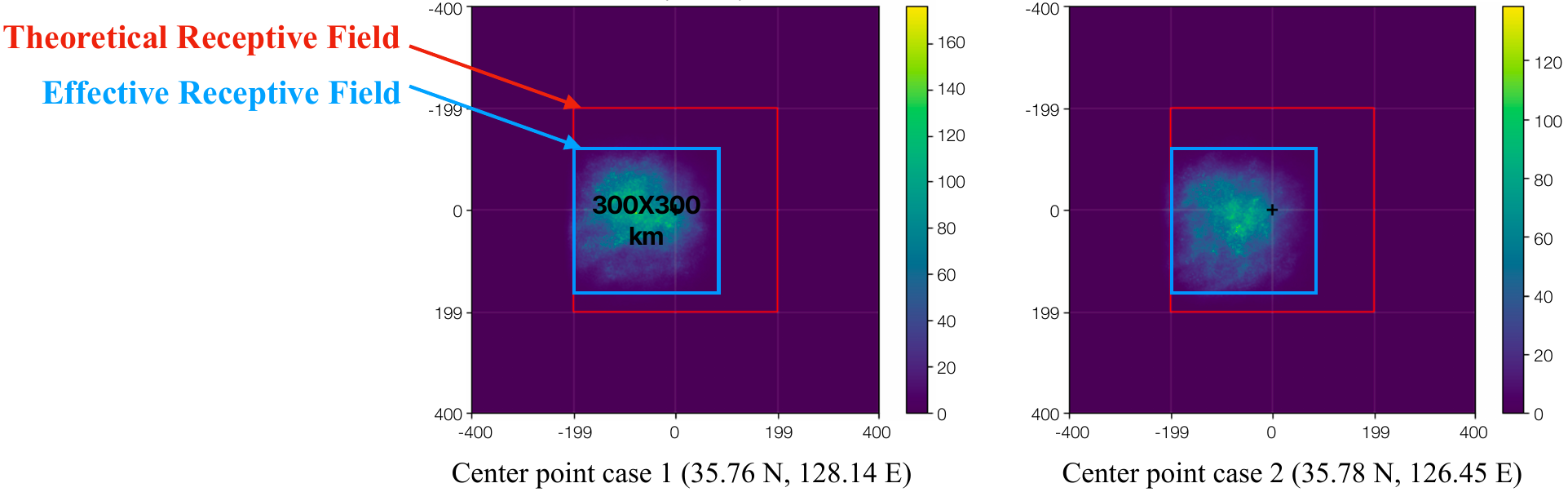}
\caption{Theoretical receptive field and effective receptive field of the target model. Due to the CNN structure, the maximum range of input region seen by a single output pixel is theoretically 398×398 km (approximately 200 km in radius). Depending on the learned parameters, the actual range is about 300×300 km (approximately 150 km in radius).}
\label{fa_receptive}
\end{figure}

\subsubsection{Qualitative Evaluation of Selected Attribution Method.}
To qualitatively evaluate the explanatory results, case-based anecdotal evidence has been analyzed through three consultations with external experts. 
Specifically, extreme precipitation cases are selected from the 2020 SOM-based classification study on the JJAS (June, July, August, and September which represent the period of the southwest monsoon) period in Korea by NIMS to match recognizable physical dynamics with attribution patterns.

The leftmost case in Figure \ref{fa_quality} is a case of the monsoon front, with a convection system moving from west to east. 
The attribution values are high at the edge of the radar area, most likely because the convective system is moving in from outside the effective range of the radar. 
This explanation can be considered an artifact.
The middle figure is an extratropical cyclone system. 
The attribution map seems to describe the disappearance signal of fragmented convection cells moving in the direction opposite to the progression of the cold front (blue line).
The rightmost case is an extratropical cyclone system.
Moist and warm air from the East Sea and the West Sea blow inland, causing friction and rising along the Taebaek Mountain Range to result in convergence. The attribution heatmap seems to concur with this phenomenon, highlighting the corresponding area.

As an additional test, the receptive fields of the target model are identified using feature attribution.
\textit{Smooth Integrated Gradient} is applied on 75 samples and the average attribution map is used for evaluating the receptive field.
As shown in Figure \ref{fa_receptive}, the effective receptive field seems to be west-biased, which aligns with the fact that the westerlies are prevalent in Korea.
The effective receptive field also has a radius of about 150 km. Assuming the maximum wind speed of 60 km per hour (about 16 m/s), the model may be making guesses when making predictions for three hours or later.

\subsection{Explanation 3: Confidence Calibration}\label{exp3}

\begin{figure}[b!]
\includegraphics[width=\textwidth]{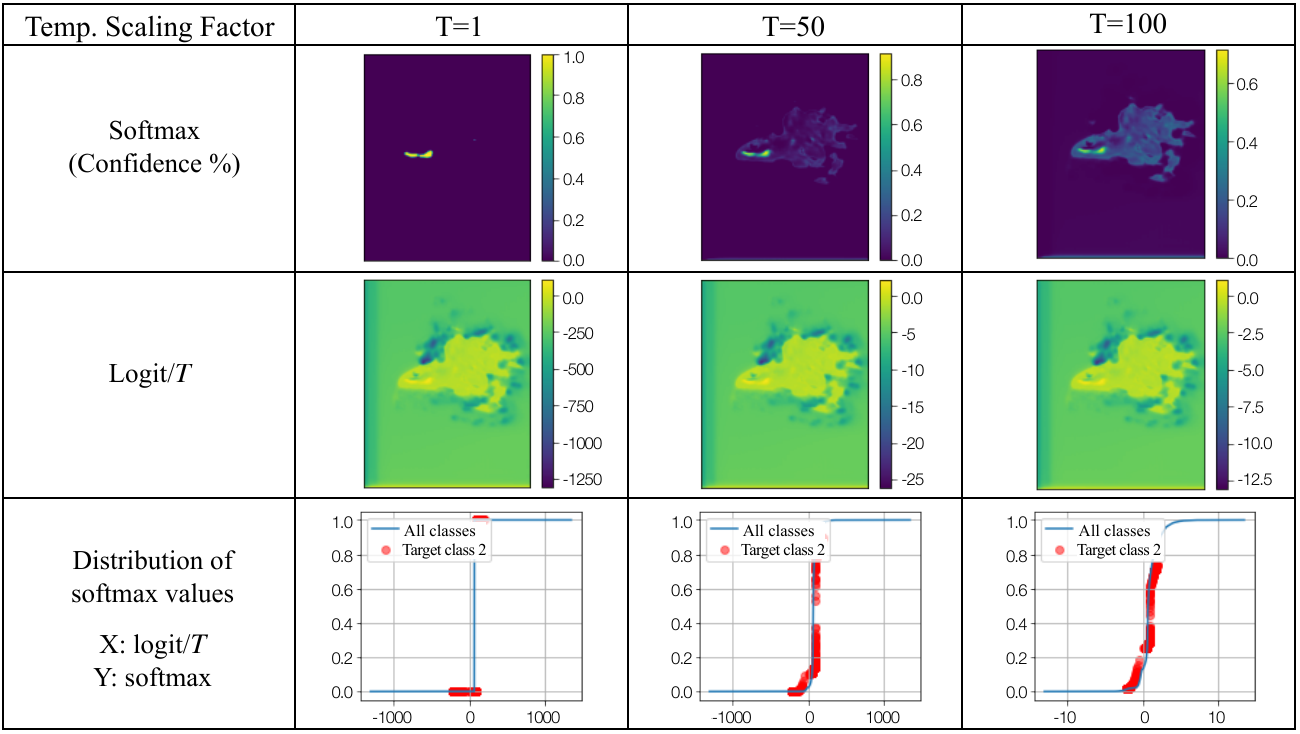}
\caption{The principle of temperature scaling. 
The softmax probability is scaled by a scalar parameter to reduce overconfidence by scaling the extreme logit values which occur near 0 or 100\% of overconfidence. 
From left to right, the probability calibration progresses.}
\label{temperature}
\end{figure}

\textit{Confidence} refers to the degree of certainty that a model has in its predictions.
The certainty can be represented as a probability, and a well-calibrated model should be capable of assigning accurate confidence probabilities to its predictions. 
Unfortunately, deep learning models trained on negative log-likelihood (NLL) tend to exhibit overconfidence since it makes low-entropy distributions of the predictive classes \cite{ding2021local} as demonstrated in Figure \ref{temperature}. 
In operational forecasting, a classification or segmentation model not only must be accurate but also indicate the point at which it is likely to be erred \cite{haynes2023creating}.
Probability calibration, the process of ensuring that the predicted probabilities of a model accurately reflect the true probabilities of the outcomes, can address this issue. \cite{guo2017calibration}

\subsubsection{Probability Calibration Methods.}
Probability calibration methods adjust the softmax of model logits as pseudo-probabilities.
This paper uses the post-processing-based probability calibration methods which do not require re-training, making it suitable for quickly adjusting large-scale weather forecasting models. 
One of the simplest non-parametric approaches is \textit{histogram binning}: all uncalibrated predictions are divided into mutually exclusive bins, enabling the selection of predictions that minimize bin-wise squared loss \cite{Zadrozny2001ObtainingCP}.
\textit{Platt scaling} is a parametric calibration method that uses a sigmoid function to calibrate non-probabilistic classification predictions for logistic regression models.
The calibrated probability $\hat{q}=\sigma(a z_i+b)$ with two parameters $a, b \in \mathbb{R}$ are optimized by NLL while model parameters are fixed \cite{platt1999probabilistic}.
\textit{Temperature scaling} (TS), on the other hand, is a variation of Platt scaling that uses a single scalar parameter $T > 0$ for all classes \cite{guo2017calibration}. 
With the logit value $z_i$ in each $i$-th pixel, the calibrated confidence is obtained as 
$\hat{q}_i (x, T) = \underset{k\in K}{\operatorname{max}}
\sigma_{SM}(z_i/T)^{(k)}.$

Where $k$ is the label index in $K$ classes and $\sigma_{SM}$ is softmax operation. 
The only learnable parameter $T$ is optimized by the NLL.
Since the maximum value of the softmax function $\sigma_{SM}$ remains unaffected by $T$, the class prediction also remains unchanged. 
This consistency of model performance makes temperature scaling suitable for the task of probability calibration.

\textit{Local temperature scaling} (LTS) \cite{ding2021local} expands on the concept of TS in semantic segmentation tasks by introducing learnable parameters for individual image pixels.
Their approach considers spatially varying temperature values and pixel-level changes. 
To achieve this, a mapping function is essential to train which takes logits $z(x)$ and the corresponding image sample $x$ as inputs and generates scaling factors $T_i(x)$. 
These scaling factors are then divided by the logits $z_i(x)$.
Formally, 
$$
\hat{q}_i (x, T_i(x))= \underset{k\in K} {\max} \sigma_{SM}(z_i(x) / T_i(x))^{(k)}
$$

where $T_i(x) \in \mathbb{R}^{+}$ is sample and pixel dependent.
We train the mapping functions for each lead time separately and employ a CNN, following a similar approach as described in the original paper.
The mapping functions are optimized by minimizing the NLL with respect to the validation dataset.

\begin{table}[t!]
\caption{Expected calibration error (ECE) of calibrated confidence on each lead time. The ECE is improved after calibration.}\label{ece}
\setlength{\tabcolsep}{1em}
\scriptsize
\centering
\begin{tabular}{ccc}
\toprule
\multirow{3}{*}{Lead Time} & \multicolumn{2}{c}{ECE} \\
\cmidrule(rl){2-3} 
 & Before & After \\
\midrule
1 hour &  0.029 & \textbf{0.010} \\
2 hour &  0.099 & \textbf{0.055} \\
3 hour &  0.170 & \textbf{0.037} \\
4 hour &  0.232 & \textbf{0.168} \\
5 hour &  0.290 & \textbf{0.109} \\
6 hour &  0.320 & \textbf{0.003} \\
\bottomrule
\end{tabular}
\label{calibration}
\end{table}

\begin{figure}[t!]
\includegraphics[width=\textwidth]{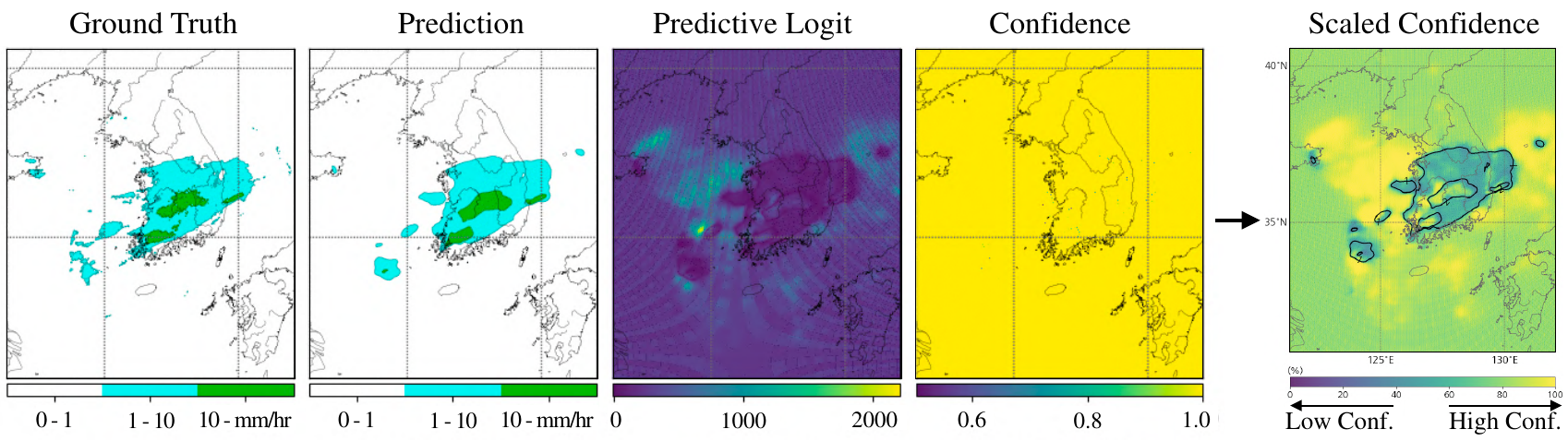}
\caption{The case of 2020-08-07 at 14:00 UTC with temperature scaling.}
\label{confidence}
\end{figure}

\subsubsection{Evaluation for Probability Calibration.}
A commonly used measure of the probability calibration of a machine learning classifier is \textit{expected calibration error} (ECE) \cite{Naeini2015ObtainingWC}. It calculates the disparity between the predicted confidence and the actual probabilities.
ECE is calculated by partitioning the range of predicted confidences into a set of bins and then calculating the weighted average discrepancy between the average confidence $\operatorname{conf}(B_i)$ and the average accuracy $\operatorname{acc}(B_i)$ within each bin $B_i$ as 
$\operatorname{ECE}=\sum_{b=1}^B\frac{n_b}{N}|\operatorname{acc}(B_i)-\operatorname{conf}(B_i)|.$

To utilize the ECE metric in the segmentation model, each pixel is considered as an individual sample as in \cite{ding2021local}.
To reduce computing costs, we randomly sample a predefined length of 250 ten times from a flattened array of confidence. 
Additionally, we masked ineffective areas in radar samples to improve the fidelity of the ECE metric by avoiding empty bins.

As shown in Table \ref{calibration}, the optimized LTS network improves the ECE scores after calibration for each of the six lead times, while maintaining the modified F1 scores. 
As demonstrated with an example in Figure \ref{confidence}, the LTS network diminishes the overconfidence in the predicted labels.
The regions of heavy rain and no rain have high confidence scores rather than those of light rain while the predictive output seems to be similar to the ground truth.

\subsection{Visualization: XAI Interface System}\label{expsys}

User interface design with XAI has been recently studied \cite{bradley2022explainable, chromik2021human}. 
In the design principles studied by Chromik et al. (2021) \cite{chromik2021human}, XAI interfaces for users should provide progressive disclosure of explanatory information in order to avoid overwhelming users. 
This can be achieved through features such as tooltips or toggle buttons. 
Additionally, considering that users are accustomed to different explanation modalities, such as natural language or visual explanations, they should be offered these modes of presentation to comprehend the information.

\begin{figure}[t!]
\includegraphics[width=\textwidth]{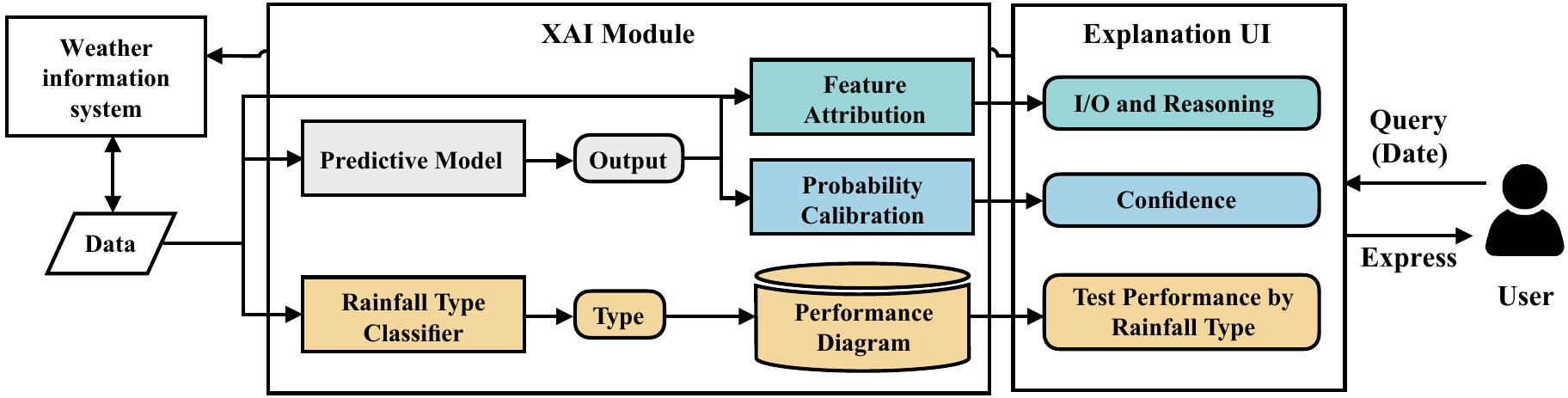}
\caption{Use case diagram for user interface and XAI modules.}
\label{usecase}
\end{figure}

\begin{figure}[b!]
\includegraphics[width=\textwidth]{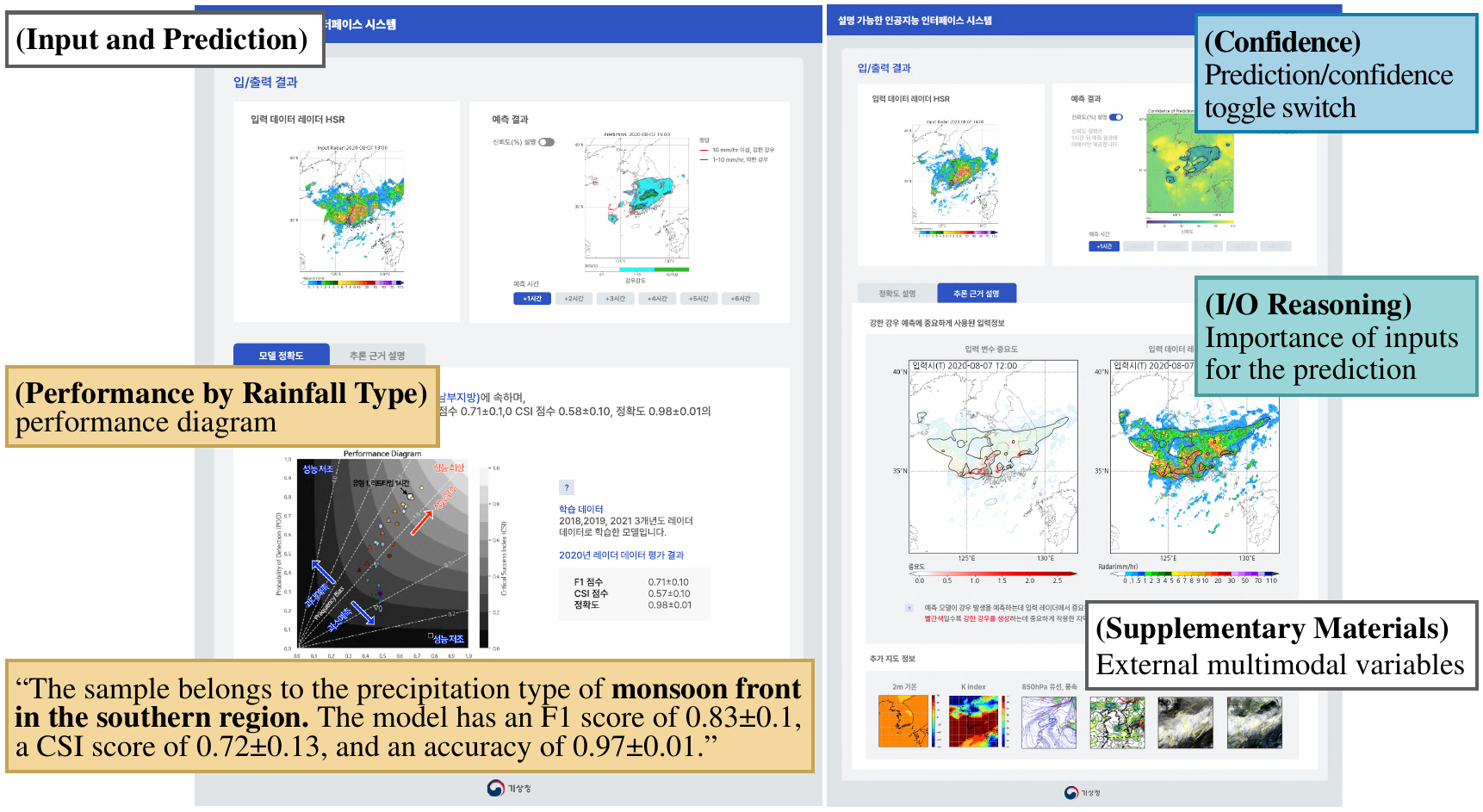}
\caption{Conceptual prototype of the interface system for the user-centered explanation. The demonstration is available (\url{https://figma.fun/LuhqIv}) in Korean}
\label{demo}
\end{figure}

In this study, a pilot interface system has been established to display the explanations in a user-friendly manner, as shown in Figure \ref{usecase} and \ref{demo}. 
The explanation components consist of four parts:
\vspace{-0.1in}
\paragraph{Performance by Rainfall Type.}
After visualizing the input and prediction, the model performance explanation panel shows the test performance for the sample's rainfall type. 
A description of the training data is also provided.
\vspace{-0.1in}
\paragraph{Output Reasoning.}
The contribution of different target classes is computed simultaneously, allowing for comparison of the input contributions to no rain, light rain, and heavy rain classes.
\vspace{-0.1in}
\paragraph{Confidence.}
To explain confidence, a toggle key is provided that allows users to compare prediction and confidence results in individual grids.
\vspace{-0.1in}
\paragraph{Supplementary Materials.}
Based on user feedback, all results are presented along with other modalities that are excluded from the model inputs. 
The additional data allows for an increase in reliability as the users can verify their opinions on the generated results.

The text and color schemes in the visuals are expressed in plain language and domain terminology.

\begin{figure}[t!]
\includegraphics[width=\textwidth]{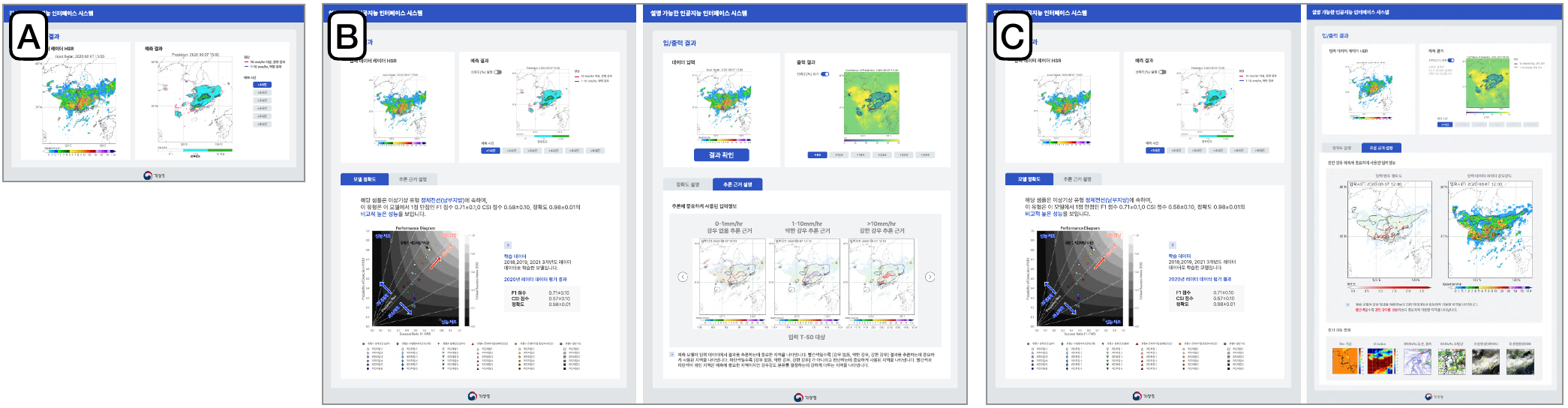}
\caption{Three prototypes for a user survey. A demonstration is available for each: prototype A providing only prediction results (\url{https://figma.fun/uQcW9P}); prototype B adding three explanatory modules (\url{https://figma.fun/6n0CgH}); and prototype C including supplemented materials from user feedback and providing the simple and contracted information in the output reasoning module (\url{https://figma.fun/LuhqIv}).}
\label{prototype}
\end{figure}

\subsubsection{User Study on XAI Interface.}
Four forecasters in Korea Meteorological Agency participated in the user study.
This user study aims to demonstrate the interface system and elicit user feedback regarding their experience. 
The purpose of this survey is to qualitatively assess whether the explanatory modules, when provided alongside the predictions of an AI model, are useful to forecasters in practice.
The survey assesses user experience based on three prototype interface systems (Q1-3) in Figure \ref{prototype} and three types of explanatory modules (Q4-6).
The participants answered 5-point Likert scale questions:
Understandability \textit{``Is the explanation easy to understand?''}, 
Usefulness \textit{``Would use in practice?''}, 
Trustworthiness \textit{``Can you trust the prediction?''}.

\begin{figure}[t!]
\includegraphics[width=\textwidth]{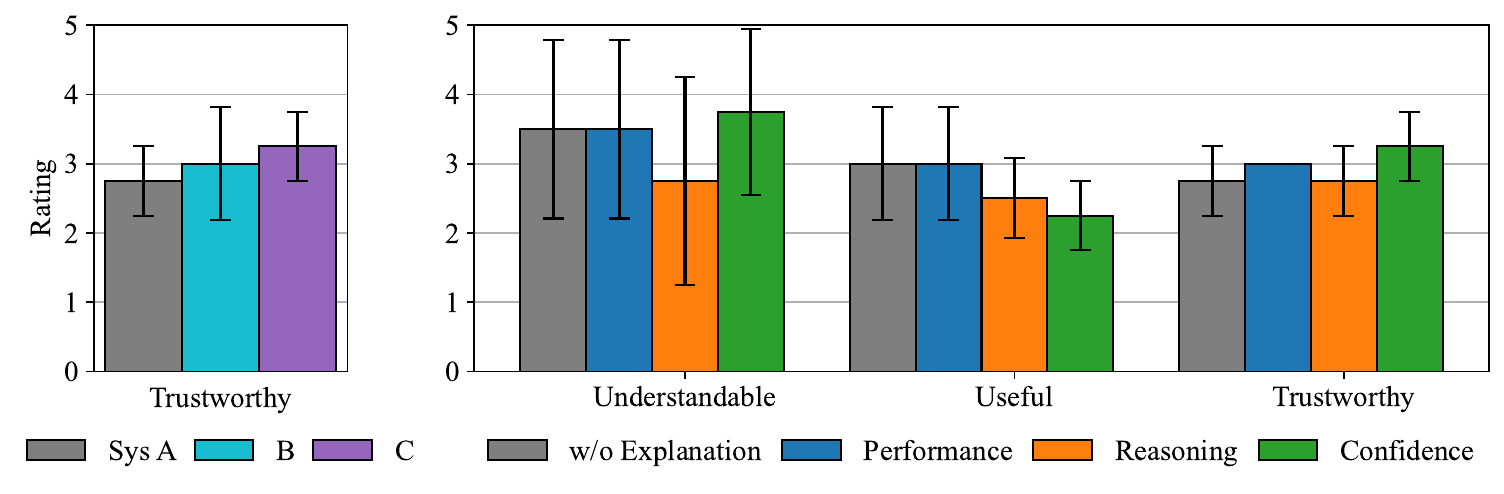}
\caption{Results from user experiment surveys. A comparison result of three prototypes of interface systems on trustworthiness (the left) and that of three explanatory modules on understandability, usefulness, and trustworthiness (the right).}
\label{survey}
\end{figure}

As results in Figure \ref{survey}, compared to no explanation system A, the users experienced more trustful in the explanatory systems B and C which provide explanatory modules (B) and simplified explanation and additional thematic maps (C), respectively.
The explanatory modules of the model performance by rainfall types (blue) and confidence (green) enhanced trustworthiness to some extent.
Unfortunately, the users found the explanations to be difficult to understand in the output reasoning explanatory module (orange).
The users also considered it unlikely to use output reasoning (orange) and confidence (green) explanatory modules in practice.
The low usefulness of these explanatory modules was induced by the effort required to understand the information since forecasters often need to make decisions quickly.
However, the participants expressed the view that for improved user acceptance and practical usability, it would be essential to establish a linkage between the XAI interface system and the existing systems employed by forecasters within the domestic meteorological agency.
Also, forecasters found the research to be promising and were receptive to the idea of further investigating AI behavior. 
This response may provide a direction for future research, focusing on XAI receptiveness from the users. 

\section{Discussion}
Through a series of meetings and interviews with the users, this study reduced the desired explanations into three main questions: model performance by rainfall type, inference reasoning, and output confidence. 
Based on the user needs and XAI algorithm mapping, a performance diagram with rainfall classifier, feature attribution, and probability calibration were selected as appropriate explanations for the requirements. 
Further analyses were performed to finalize the specific XAI methods in each category. 
Finally, three prototypes of the user interface were designed and feedback is received from the users.
User experience survey results of the explanatory modules were promising on trustworthiness.
Forecasters, however, requested high standards for actual use in practice since forecasters commonly need rapid decision-making.

One limitation of this study is that the overall process involved a specific set of users in Section \ref{exp0}; hence, the results may not cover the entirety of possible user requirements, creating a gap between XAI results and individual users' desired approaches as discussed in \cite{liao2021human}. 

Another limitation is that for model performance by rainfall type in Section \ref{exp1}, the classifier shows limited performance due to a lack of samples with rainfall type labels. 
For actual implementation, it would be necessary to train the classifier with a larger dataset.

While the feature attribution methods in Section \ref{exp2} can faithfully reflect model reasoning, even for distinct rainfall types, it can be challenging for experts to interpret. 
One reason for this difficulty is the model's reliance on uni-modal input features, restricting the feature attribution results to highlighting only the horizontal movement of convection cells.
This issue may be addressed by using multi-modal data -- in particular, since radar observations only represent the final outcomes of various physical mechanisms and the radar product used for training the target model provides only horizontal information, it would be ideal to include additional features that can provide this information.


In Section \ref{expsys}, users have provided feedback that the feature attribution explanations are hard to understand even if the explanations show high fidelity. 
This opinion indicates the need to measure the complexity of explanation results. 
Thus, user-centric XAI performance may need to reflect qualities of explanation besides faithfulness.
Several previous works use Shannon entropy to measure complexity in the image domain \cite{bhatt2020evaluating}, but it is essential to recognize that the proxy variables in the weather domain have different characteristics due to their spatiotemporal context and may require different metrics of complexity.

Our pilot interface system is clickable, but it is a shallow-level user-interactive XAI (UXAI) system that becomes static after the completion of user-centric building procedures.
Providing high-level interaction makes a potential area for future work to support explanations in response to feedback from the users such as interactive dialogue \cite{chromik2021human}.

\section{Conclusion}
This study emphasizes the significance of involving users as key stakeholders in the design process of explainable artificial intelligence (XAI) systems. 
Based on an analysis of user requirements in the meteorological domain and the mapping of these requirements to XAI methods, rainfall classification, feature attribution, and probability calibration are selected as suitable explanations. 
By presenting the model's performance for each rainfall type, users can judge the overall reliability of the corresponding AI model.
Furthermore, sample cases of alignment with domain knowledge for feature attribution are identified. 
This investigation helps determine the practical applicability of feature attribution methods in meteorology.
By providing confidence explanations for each output grid, users can assess the likelihood of output accuracy and decide the local reliability of individual predictions.
Lastly, three prototypes of the user interface are designed and solicited feedback from users to ascertain the feasibility of integrating XAI into the forecasting system.
This study may contribute to the literature as a use case of user-centered expression research.



%
%

%




\end{document}